\documentclass[10pt,twocolumn,letterpaper]{article}

\usepackage[pagenumbers]{cvpr} 

\usepackage[textsize=tiny]{todonotes}
\definecolor{myblue}{HTML}{1f78b4}
\definecolor{mybluelight}{HTML}{a6cee3}

\definecolor{mygreen}{HTML}{fb9a99}
\definecolor{mygreenlight}{HTML}{33a02c}

\definecolor{mypurple}{HTML}{6a3d9a}
\definecolor{mypurplelight}{HTML}{cab2d6}

\definecolor{myred}{HTML}{fdbf6f}
\definecolor{myredlight}{HTML}{e31a1c}

\definecolor{myolive}{HTML}{ada136}
\definecolor{myolivelight}{HTML}{7ead36}

\renewcommand{\paragraph}[1]{\vspace{.5em}\noindent\textbf{#1.}}

\definecolor{cvprblue}{rgb}{0.21,0.49,0.74}
\usepackage[pagebackref,breaklinks,colorlinks,allcolors=cvprblue]{hyperref}
\usepackage{stfloats} 
\def\paperID{22} 
\def\confName{CVPR}
\def\confYear{2026}

\title{EMO-BOOST: Emotion-Augmented Audio-Visual Features for Improved Generalization in Deepfake Detection}

\author{Aritra Marik$^{1,2}$\quad
Marcel Klemt$^{1,3}$\quad
Anna Rohrbach$^{1,3}$\\[4pt]
$^1$Technical University of Darmstadt \quad $^2$ELIZA \quad $^3$hessian.AI \\[2pt]
{\tt\small aritra.marik@stud.tu-darmstadt.de} \quad
{\tt\small marcel.klemt@tu-darmstadt.de}
}
\begin{document}
\maketitle
\begin{abstract}
With every advancement in generative AI models, forensics is under increasing pressure. The constant emergence of new generation techniques makes it impossible to collect data for each manipulation to train a deepfake detection model. Thus, generalizing to deepfakes unseen during training is one of the major challenges in current deepfake detection research. To tackle this challenge, we employ high-level semantic cues and argue that these cues can support low-level focused approaches in generalizing to unseen types of manipulations. In this work, we study emotions as a high-level semantic cue. We propose Emo-Boost, a multimodal deepfake detection framework that fuses an off-the-shelf RGB- and acoustic-focused deepfake detector with our emotion-based deepfake detector EmoForensics. EmoForensics utilises vision and audio emotion recognition modules and models intra- and inter-modal temporal consistency in emotion representations from an audio-visual stream. We found that EmoForensics and the low-level focused method capture complementary signals. Consequently, combining both signals in EmoBoost enhances the average cross-manipulation generalization AUC by 2.1\% on FakeAVCeleb.
\end{abstract}    
\section{Introduction}
\label{sec:intro}

Generative AI is evolving faster than ever before. This enables the creation of images, videos, and speech for personal enjoyment and for commercial purposes. At the same time, however, it also carries the risk of \textit{deepfakes}. Deepfakes are media that convincingly imitate real individuals. In this work, we focus on the visual and auditory modalities. They are becoming increasingly difficult for us humans to detect as generation techniques advance. Yet, they pose serious societal risks, including fraud, misinformation, and identity manipulation~\cite{Chen_Magramo_2024, Loftus_2025, Catil_2025}.
Consequently, developing reliable deepfake detection methods is an ever-more pressing research challenge. Since convincing deepfakes affect not only images but also videos and audio, as well as combinations thereof, current research has shifted to multimodal deepfake detection.

\begin{figure}[t]
  \centering
   \includegraphics[width=0.835\linewidth]{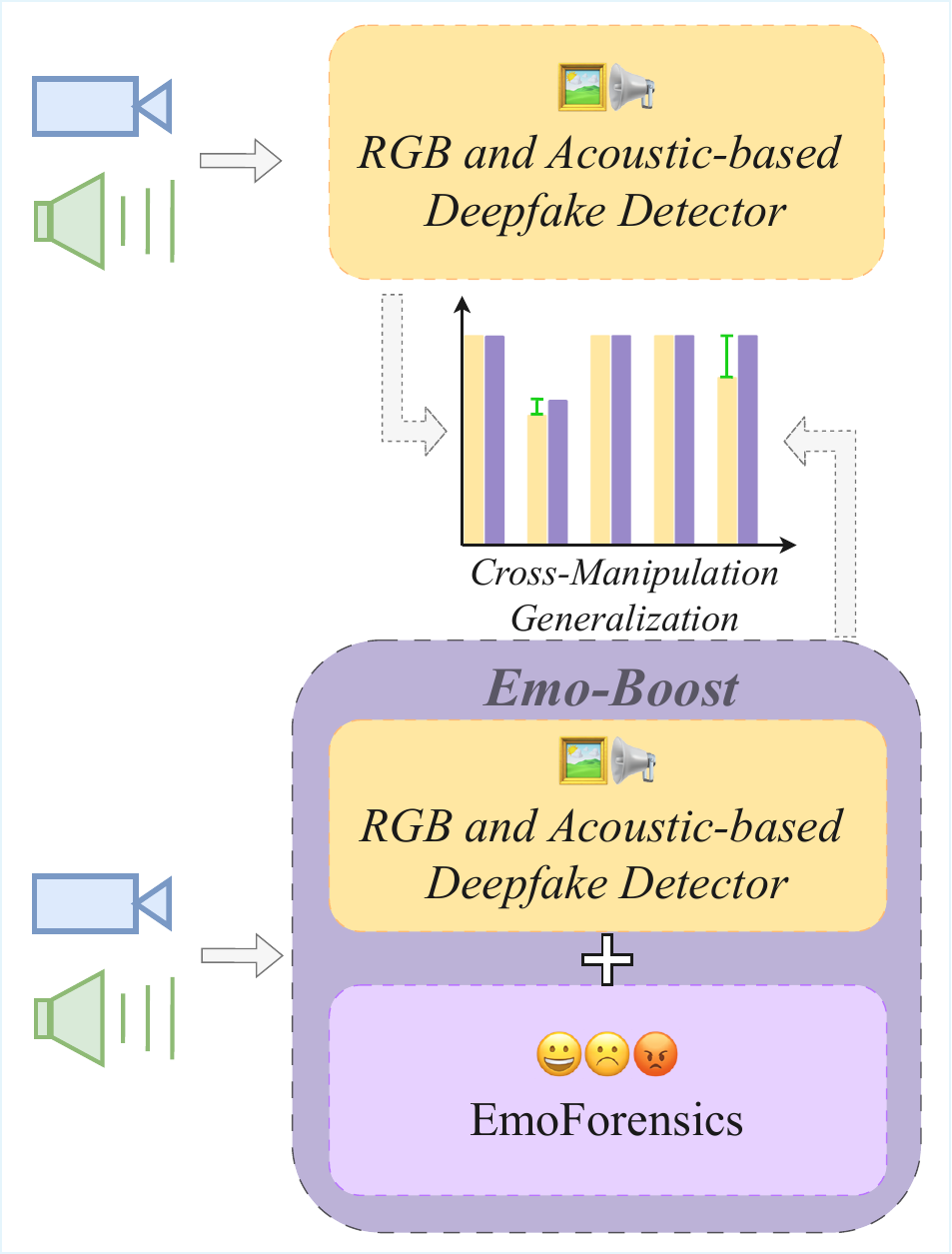}
   \caption{High-level illustration of \textbf{Emo-Boost}, a multimodal deepfake detection framework that combines an off-the-shelf RGB- and acoustic-based detector with \textbf{EmoForensics}, our emotion-based deepfake detection network. Emo-Boost pipeline (purple) improves cross-manipulation generalization compared to the standalone RGB- and acoustic-based baseline (yellow).}
   \label{fig:onecol}
\end{figure}

Multimodal deepfake detection jointly analyses visual and audio cues and exploits unimodal signal as well as cross-modal inconsistencies such as phoneme–viseme misalignment and temporal desynchronization~\cite{yang2023avoid, oorloff2024avff, raza2023multimodaltrace, liu2024lips, klemt2025deepfakedoctordiagnosingtreating}. These models can detect manipulations stemming from both modalities, the visual and the auditory. However, most existing methods primarily rely on low-level spatial, temporal, or spectral features extracted from raw data. In contrast, higher-level semantic cues remain underexplored in this domain.

In particular, affective information has received limited attention in deepfake detection, although deepfake generation techniques might struggle to display affective information convincingly~\cite{lopez-gil}. Human emotion represents a rich signal reflecting natural facial and vocal expressivity, and inconsistencies in affective behavior may provide useful cues for identifying manipulated media. A small number of studies have begun exploring this direction. For example, \textit{Emotions Don’t Lie}~\cite{mittal2020emotions} classifies real and fake videos using emotion embeddings together with spatiotemporal features in a Siamese configuration,  while \textit{Do DeepFakes Feel Emotions?}~\cite{hosler2021deepfakes} focuses only on emotion-based artefacts and models temporal affective dynamics across modalities to detect manipulated videos. Although these studies leverage emotion representations as a cue for deepfake detection, they do not explicitly account for the distribution shift in the emotion recognition networks introduced by manipulated media. Furthermore, multimodal detection methods often struggle with in-domain generalization evaluation, as shown in~\cite{klemt2025deepfakedoctordiagnosingtreating}. We hypothesize that higher-level semantic signals, particularly those derived from emotion features, may provide more stable cues that help these existing methods generalize beyond manipulation-specific artefacts.

In this work, we propose \textbf{EmoBoost} (\Cref{fig:onecol}), a framework that enhances RGB- and acoustic-focused deepfake detectors with emotion-aware representations. For our audio-visual emotion-based deepfake detector, we introduce \textbf{EmoForensics}. EmoForensics is designed along two dimensions for modeling emotion representations: i) \emph{Intramodal Temporal Consistency in Emotion representations} to capture the limitations of deepfake generators to produce consistent natural emotion expressions over time within a particular modality, and ii) \emph{Intermodal Temporal Consistency in Emotion representations} to capture inconsistent emotion expressions in the synthesized visual and audio stream. We leverage two frozen emotion-detection encoders to extract emotion representations from the visual and audio stream~\cite{zheng2023poster, ma2023emotion2vec}. We follow the emotion encoders with modality-specific transformer encoders to model intramodal temporal consistency of emotion representations. To encourage the model to capture intermodal consistency between audio and visual emotion dynamics, we employ contrastive loss in addition to binary cross-entropy. 
To summarise, our contributions are as follows, 
\begin{itemize}
    \item We propose an emotion-augmented audio-visual deepfake detection framework \textbf{Emo-Boost}, that integrates an emotion-based deepfake detection network with an off-the-shelf multimodal deepfake detection method using simple late-stage fusion. 
    \item We introduce a multimodal emotion-based deepfake detector \textbf{EmoForensics}, which models two complementary dimensions: (i) intramodal temporal consistency of emotion representations, and (ii) intermodal temporal consistency of emotion representations from the audio and visual streams. 
    \item Emo-Boost shows promising improvements in deepfake detection performance when evaluated in a cross-manipulation evaluation scenario. The average AUC increases, for example, by 2.1\% on the popular FakeAVCeleb~\cite{khalid2021fakeavceleb} benchmark compared to the respective low-level detector. On DeepSpeak v2~\cite{barrington2024deepspeak}, Emo-Boost achieves performance competitive with existing state-of-the-art methods.
\end{itemize}

\section{Related Works}
\label{sec:rel_works}

\subsection{Emotion Recognition}

Facial emotion recognition (FER) aims to infer affective states from facial signals and is a central problem in affective computing. Static FER (SFER), which operates on images, has progressed significantly with large in-the-wild datasets such as FER-2013~\cite{goodfellow2013challenges} and AffectNet~\cite{mollahosseini2017affectnet}. These benchmarks enabled the transition from handcrafted features to deep CNN and Transformer-based models that are robust to pose, illumination, and occlusion. Representative approaches include CNN-based pipelines~\cite{cai2018island, zhang2020identity}, attention-augmented models~\cite{xue2021transfer}, and recent ViT/MAE variants that learn fine-grained facial semantics~\cite{li2022affective}

Dynamic FER (DFER) extends this setting to video and introduces challenges related to temporal modeling, emotion intensity variation, and sparse supervision. Benchmarks such as DFEW~\cite{jiang2020dfew}, FERV39K~\cite{wang2022ferv39k}, and MAFW~\cite{liu2022mafw} provide in-the-wild clips with spontaneous affective dynamics. Methodologically, recent works explore self-supervised video representation learning (e.g.\ MAE-DFER~\cite{sun2023mae}), static-to-dynamic transfer via temporal adapters (S2D~\cite{chen2024staticold}), multimodal fusion strategies (MMA-DFER~\cite{chumachenko2024mma}), and unified static-dynamic pretraining frameworks such as S4D~\cite{chen2024static}. Other lines include CNN-RNN hybrids~\cite{liu2020saanet, yu2018spatio}, transformer-based temporal models~\cite{li2016ternary, liu2023expression}, and intensity-aware sampling approaches~\cite{li2023intensity}.

Beyond vision, Speech Emotion Recognition (SER) models acoustic cues such as pitch and spectral dynamics. Earlier methods rely on handcrafted features combined with CNN or RNN architectures~\cite{pepino2021emotion, morais2022speech}, while recent approaches adapt large self-supervised speech encoders such as wav2vec2.0, HuBERT, and WavLM~\cite{wang2021fine, ioannides2023towards}. Emotion2vec~\cite{ma2023emotion2vec} further proposes a universal speech emotion representation learned via self-supervised distillation. Together, advances in FER and SER suggest that emotion representations capture semantically rich and temporally grounded signals that may be informative beyond recognition alone.

\subsection{Deepfake Detection}

\paragraph{Unimodal Detection}
Unimodal deepfake detectors typically operate on either video or audio. Visual methods range from CNN baselines such as XceptionNet~\cite{rossler2019faceforensics++} to approaches that exploit facial geometry, head-pose inconsistencies~\cite{yang2019exposing}, or region-specific cues, such as lip motion (LipForensics~\cite{haliassos2021lips}). Hybrid CNN--Transformer models capture both spatial artefacts and temporal incoherence~\cite{zheng2021exploring, wodajo2021deepfake}, while recent studies address generalization and identity leakage~\cite{ojha2023towards, dong2023implicit}.  
Audio-only detectors analyze speech spectrograms using CNNs or capsule networks to capture synthetic speech patterns~\cite{wani2023deepfakes, wani2024abc}. While effective in constrained settings, unimodal methods struggle against modern forgeries that manipulate multiple modalities.

\paragraph{Multimodal Detection}
Multimodal approaches jointly analyze audio and video, often targeting cross-modal inconsistencies. Strategies include phoneme--viseme mismatch detection~\cite{agarwal2020detecting}, identity coherence modeling~\cite{tian2023unsupervised}, and joint audio-visual representation learning~\cite{cheng2023voice, zhou2021joint}.  
Recent methods such as AVoiD-DF~\cite{yang2023avoid}, AVFF~\cite{oorloff2024avff}, and MultimodalTrace~\cite{raza2023multimodaltrace} employ self-supervised or contrastive objectives to learn aligned audio-visual embeddings. Other works focus on specific cues, including lip synchronization~\cite{liu2024lips} or temporal audio-visual alignment anomalies~\cite{feng2023self}. These methods highlight the effectiveness of exploiting cross-modal disharmony in detecting sophisticated deepfakes.

\subsection{Emotion-Based Deepfake Detection}

\textit{Emotions Don't Lie}~\cite{mittal2020emotions} was one of the earliest papers to investigate the role of emotions in deepfake detection. The authors use both generic modality and perceived emotion representations with a CNN-based network and a Memory-Fused Network (MFN)~\cite{zadeh2018memory}, respectively. The model is trained in a Siamese configuration with a contrastive objective that enforces high similarity between the face and speech embeddings of real (non-manipulated) clips while discouraging such agreement when one modality has been manipulated. During inference, distances between the face-speech modality embeddings and their corresponding emotion embeddings are combined and compared against a learned threshold to classify videos as real or fake.

\textit{Do Deepfakes feel Emotions?}~\cite{hosler2021deepfakes} uses affective signals as a cue for multimodal deepfake detection by modeling temporal valence-arousal trajectories from facial action units and acoustic features, detecting forgeries via cross-modal and temporal emotion coherence. From the resulting temporal emotion curves, the authors explore two detection strategies: (i) a feature-based approach using simple statistical summaries of cross-modal emotion signals and (ii) a sequence model that feeds the four emotion time series directly to a lightweight bidirectional LSTM classifier. By focusing on cross-modal and temporal coherence in perceived affect, the approach aims to identify manipulations without relying on pixel-level artefacts or specific generation techniques.

Additional studies explore emotion-based speech deepfake detection~\cite{contiDeepfake} or analyze how well deepfakes preserve facial affect without proposing detectors~\cite{lopez-gil}. While these works suggest that affective inconsistencies are indicative of manipulation, they are typically evaluated on limited datasets and often train their pretrained emotion recognition models on deepfake datasets~\cite{mittal2020emotions, hosler2021deepfakes}. As a result, emotion recognition models may implicitly learn affective patterns present in the manipulated videos. This mitigates the distribution shift, which could potentially be caused by the generative models, thus resulting in the lack of investigation of out-of-distribution behavior of emotion recognition models when applied to manipulated videos.

In contrast, we propose \textbf{Emo-Boost}, a framework designed to enhance existing RGB- and acoustic-based deepfake detector models by integrating complementary emotion-based signals. We integrate emotion-based signals with an off-the-shelf deepfake detector via simple element-wise multiplication, improving their in-domain generalization performance by a significant margin. To provide these complementary signals, we propose \textbf{EmoForensics}, which extracts emotion representations using pretrained emotion encoders that account for manipulation-induced distribution shifts. Beyond emotion embeddings, it further models both intra- and intermodal temporal consistency in emotion representations, enabling the framework to capture affective inconsistencies introduced by manipulated media.

Unlike \textit{Emotions Don't Lie}~\cite{mittal2020emotions}, Emo-Boost accounts for manipulation-induced distribution shifts in pretrained emotion encoders and leverages temporal consistency in emotion representations rather than relying on cross-modal similarity alone. In contrast to \textit{Do Deepfakes Feel Emotions?}~\cite{hosler2021deepfakes}, Emo-Boost uses emotion signals as complementary cues fused with RGB and acoustic features, while also addressing distribution shifts introduced by deepfake manipulations.

\section{Methodology}
\label{sec:methodology}

\begin{figure*}[htbp]
    \centering
        \includegraphics[width=1.0\textwidth]{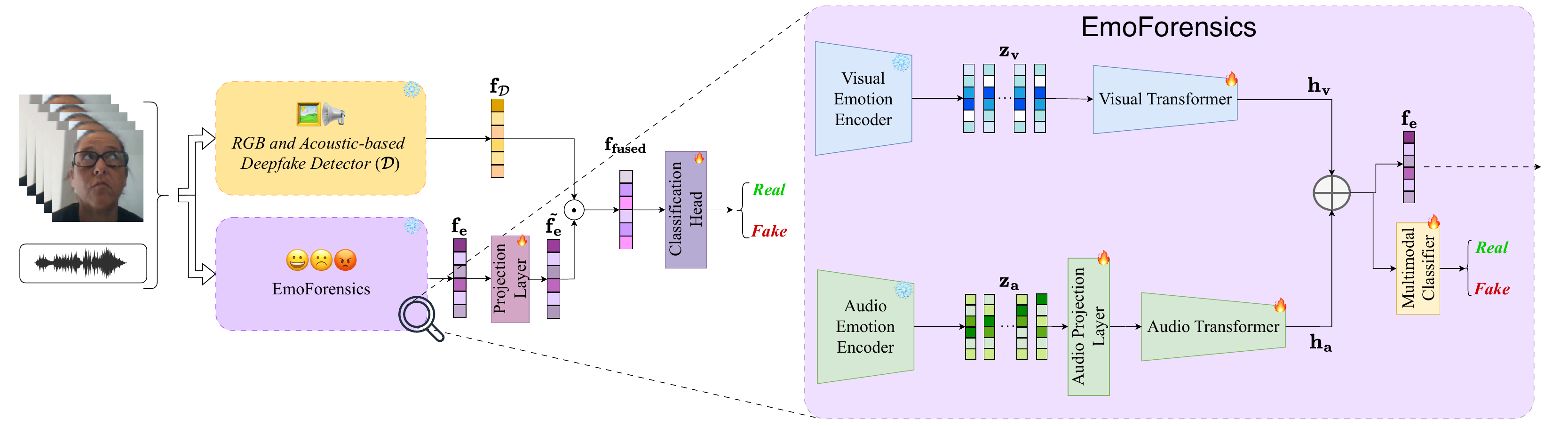}
    \caption{Overview of our proposed framework, \textbf{Emo-Boost}, and the emotion-based deepfake detection network, \textbf{EmoForensics}. Given a video and audio input, frozen emotion encoders produce frame-level visual and audio representations, $\mathbf{z_v}$ and $\mathbf{z_a}$. These frame-level embeddings are passed through their respective modality-specific transformers to obtain modality-level emotion representations, $\mathbf{h_v}$ and $\mathbf{h_a}$. The modality-level representations are fused via element-wise addition to form $\mathbf{f}_e$, which is fed to the multimodal classifier. For Emo-Boost, $\tilde{\mathbf{f_e}}$ (a projected version of $\mathbf{f_e}$) is combined with the feature representation $\mathbf{f}_{\mathcal{D}}$ from an off-the-shelf RGB- and acoustic-based deepfake detector $\mathcal{D}$, producing the fused representation $\mathbf{f}_\text{fused}$ that is fed to the final classification head. EmoForensics is trained independently, and in Emo-Boost, both the trained EmoForensics and the pre-trained low-level detector $\mathcal{D}$ are kept frozen.}
    \label{fig:emo-boost-pipeline}
\end{figure*}

\subsection{EmoForensics}

We propose \textbf{EmoForensics}, a deepfake detection framework that models emotion-representation-based signals from audio and visual streams. The overall architecture is illustrated in~\Cref{fig:emo-boost-pipeline}. Our method is motivated by the hypothesis that, in deepfake media, emotion expression is disrupted across a single modality or both modalities, thereby causing a manipulation-induced distribution shift in emotion recognition networks trained on real videos. To capture such artefacts, we model emotion representations along two complementary dimensions: First, we expect \textit{intramodal temporal consistency of emotion representations} to be lower for deepfakes, as it is a hard problem to synthesize a video that realistically shows a consistent emotion over time. Second, we utilize intermodal temporal consistency between visual and acoustic emotion representations. Generating a consistent emotion across both modalities is assumed to be an even harder task, especially when only one modality is modified.

\paragraph{Emotion Representation extraction}
Rather than learning emotion features from scratch, we use pretrained emotion encoders to obtain frame-wise emotion embeddings from both modalities. For the visual stream, we employ a frozen facial emotion recognition model as the visual encoder to produce frame-level emotion representations. Given a sequence of $T_v$ frames from a video $\mathbf{x}$, the encoder generates a sequence of emotion embeddings $\mathbf{z_v} \in \mathbb{R}^{T_v \times d_v}$, where $d_v$ denotes the embedding dimensionality. Similarly, for the audio stream, we use a frozen speech emotion recognition model as the audio encoder to extract frame-level speech emotion representations. The audio signal is processed to produce frame-level embeddings of dimension $d_a$, resulting in $\mathbf{z_a} \in \mathbb{R}^{T_a \times d_a}$. These pretrained encoders remain frozen during training, allowing EmoForensics to exploit emotion representations learned from large-scale emotion datasets while avoiding overfitting to manipulation distribution shifts.

\paragraph{Intra- and Inter-Modal Temporal Modeling}
For the frame-level embeddings from the visual and audio encoders, we employ two transformer pipelines based on the Temporal Transformer proposed in MMA-DFER~\cite{chumachenko2024mma}. The visual emotion representations $\mathbf{z_v} \in \mathbb{R}^{T_v \times d_v}$ are used directly. In contrast, the audio emotion representations $\mathbf{z_a} \in \mathbb{R}^{T_a \times d_a}$ are first projected into a shared feature space using a linear projection layer to align their dimensionality with that of the visual emotion representations. 

In EmoForensics, we apply temporal transformer encoders for both visual and audio modalities. Each transformer layer consists of multi-head self-attention followed by a feed-forward network, allowing the model to capture temporal dependencies within each modality. Positional encodings are added to the token embeddings to preserve the temporal ordering of the sequence. The output corresponding to the classification token from the final transformer layer provides a compact modality-level representation, denoted as $\mathbf{h_v}$ for video and $\mathbf{h_a}$ for audio.

\paragraph{Deepfake Classification}
The modality-specific representations $\mathbf{h_v}$ and $\mathbf{h_a}$ are fused using element-wise addition to obtain a joint representation
$\mathbf{f_e} = \mathbf{h_v} + \mathbf{h_a}$,
which is subsequently processed by a final linear layer classification head to predict whether the input video is real or manipulated.

\paragraph{Learning Objective}
The network is trained using a combination of binary cross-entropy (BCE) loss and contrastive loss. The BCE loss supervises the final classification head with ground-truth label $y \in \{0,1\}$:
\begin{equation}
    \mathcal{L}_{\text{BCE}} = - \left( y \log(\hat{y}) + (1-y)\log(1-\hat{y}) \right),
\label{eq:bce_loss}
\end{equation}
where $\hat{y}$ denotes the predicted probability of the video being manipulated.

To learn the consistency across modalities in authentic samples while separating manipulated pairs, we introduce a contrastive loss between the normalized video and audio representations. Within a training batch, video and audio embeddings originating from the same real video are treated as positive pairs and are encouraged to lie close in the embedding space. Negative pairs are constructed by pairing embeddings from manipulated samples with embeddings from real samples (i.e., fake–video/real–audio and real–video/fake–audio combinations), encouraging the model to push emotion representations generated from manipulated videos further apart. We do not consider fake–fake pairs, as the objective is to explicitly separate real emotion representations from manipulated ones rather than enforce structure among manipulated samples. Let $d(\mathbf{h_v}, \mathbf{h_a})$ denote the Euclidean distance between the normalized embeddings. For a pair label $y_c \in \{0,1\}$ indicating whether the pair corresponds to a real (positive) or manipulated (negative) pairing, the contrastive objective is defined as
\begin{equation}
\mathcal{L}_{\text{contrast}} =
y_c \, d(\mathbf{h_v}, \mathbf{h_a})^2 +
(1-y_c) \, \max(0, m - d(\mathbf{h_v}, \mathbf{h_a}))^2,
\label{eq:contrast_loss}
\end{equation}
where $m$ is a margin parameter.

The final training objective combines the two losses:
\begin{equation}
    \mathcal{L} = (1-\alpha)\mathcal{L}_{\text{BCE}} + \alpha \mathcal{L}_{\text{contrast}},
\label{eq:comb_loss}
\end{equation}
where $\alpha$ is an experimentally chosen hyperparameter to control the relative contribution of the contrastive and classification objectives.

\subsection{Emo-Boost}

In the proposed \textbf{Emo-Boost} framework, we use EmoForensics as an add-on network to further complement the learned representations of an RGB and acoustic-based multimodal deepfake detection network with its emotion-based knowledge. For generality, we denote the off-the-shelf multi-modal deepfake detection network as $\mathcal{D}$. It is important to note that EmoForensics and $\mathcal{D}$ are trained independently and remain frozen while training the Emo-Boost framework. 

Given an input video, both EmoForensics and $\mathcal{D}$ process the same frames from a video $\mathbf{x}$  and output their corresponding feature representation. Let $\mathbf{f_e} \in \mathbb{R}^{d_e}$ denote the emotion representation obtained from EmoForensics and $\mathbf{f_\mathcal{D}} \in \mathbb{R}^{d_\mathcal{D}}$ denote the feature representation from $\mathcal{D}$.
To align the feature dimensions, we introduce a trainable MLP-based projection head that maps the EmoForensics representation to the same embedding dimension as $\mathcal{D}$. The projected EmoForensics features $\tilde{\mathbf{f_e}}$ are then fused with the $\mathcal{D}$ features $\mathbf{f_\mathcal{D}}$ using element-wise multiplication: 

\begin{equation}
    \mathbf{f_{fused}} = \tilde{\mathbf{f_e}} \odot \mathbf{f_\mathcal{D}},
    \label{eq:elem_mult_fusion}
\end{equation}

The resultant fused representation, $\mathbf{f_{fused}}$, is passed through a trainable single-layer classification head to predict whether the input video is real or manipulated. The Emo-Boost framework is trained using a binary cross-entropy loss applied to the final prediction, as defined in~\Cref{eq:bce_loss}.

The only trainable components of the Emo-Boost framework are the projection and classification heads, while the EmoForensics pipeline and $\mathcal{D}$ remain frozen. This design choice demonstrates that the proposed EmoForensics method can augment the performance of off-the-shelf RGB and acoustic-based deepfake detectors without requiring joint training.
\section{Experimental Setup}
\label{sec:exp_setup}

\paragraph{Datasets and Data Preprocessing}
We evaluate our approach on the established \textbf{FakeAVCeleb}~\cite{khalid2021fakeavceleb} and recent \textbf{DeepSpeak v2}~\cite{barrington2024deepspeak} dataset. FakeAVCeleb comprises 21k videos from 500 identities, with manipulations generated using three different video and one audio generation techniques. The more recent DeepSpeak v2 includes 280 identities and 16.5k videos with greater appearance variety, including occlusions, than the established dataset. Fakes are created using six video and four audio manipulations.

Following~\cite{klemt2025deepfakedoctordiagnosingtreating, oorloff2024avff}, we extract the face region from each video and sample $16$ frames with a step size of $5$ per video clip. For the corresponding audio stream, the EmoForensics pipeline processes the entire audio to produce frame-level embeddings through the audio emotion encoder. These frame-level audio embeddings are subsequently downsampled to the temporal length of the frame-level visual embeddings before being processed by the respective modality-specific transformer pipelines.

\paragraph{Evaluation Setup and Metrics} We evaluate our methods on two experimental settings. For the \textbf{in-domain} evaluation, we split the dataset at a standard 70\%-30\% train-test ratio. In-domain means the model has seen all manipulation types it is evaluated on during training. The \textbf{Leave-one-out (Cross-manipulation) Evaluation} tests the model's ability to classify manipulation types that were not present during training. For FakeAVCeleb, we follow the evaluation protocol described in~\cite{klemt2025deepfakedoctordiagnosingtreating}, where individual or even families of manipulations are left out during training and performance is reported on the left-out splits. A similar strategy is applied for the newer DeepSpeak v2 benchmark. Since the objective of the cross-manipulation setting is to evaluate generalization to unseen manipulations, the standard in-domain validation split cannot be used for model selection. Instead, we construct a dedicated \textit{val-test} split to validate design choices and tune hyperparameters for Emo-Boost. For each cross-manipulation setting, this split is created by uniformly sampling 20\% of both real and fake videos from the corresponding test set. We report the ablation results on this split in~\Cref{sec:results}. Detailed descriptions of the dataset splits and manipulation-specific partitions for both datasets are provided in the Appendix. For both evaluation setups, the corresponding EmoForensics and Emo-Boost instances are trained on the same dataset. 

We report the performance of our methods using the standard metrics: Average Precision (AP) and Area Under the ROC Curve (AUC). AP measures the precision of predictions with higher values indicating fewer false positives, while AUC measures the model's ability to distinguish between positive and negative classes across all decision thresholds.

\paragraph{Implementation Details}~For EmoForensics, we use POSTER~\cite{zheng2023poster} as the visual emotion encoder and emotion2vec~\cite{ma2023emotion2vec} as the audio emotion encoder. These pretrained models produce frame-level embeddings of dimensionality $d_v=512$ and $d_a=1024$ for the visual and audio streams, respectively. The frame-wise embeddings are subsequently processed using modality-specific 2-layer temporal transformer pipelines (with the depth selected experimentally), producing a final $512$-dimensional emotion representation.
The EmoForensics model is trained for 100 epochs using the AdamW optimizer with an initial learning rate of $1e-3$, weight decay of $0.05$, and $\epsilon = 1e-8$. We apply dropout with a rate of $0.15$ and use a Reduce-on-Plateau learning rate scheduler with a patience of 4 epochs. Early stopping is applied with a patience of 50 epochs. We set the $\alpha$ hyperparameter in the final learning objective of EmoForensics to $0.5$.

For Emo-Boost, we use SIMBA~\cite{klemt2025deepfakedoctordiagnosingtreating} as the off-the-shelf multimodal deepfake detector. Emo-Boost is trained for 20 epochs with an initial learning rate of $1e-3$, using the same optimizer and learning rate scheduler as in EmoForensics, and early stopping with a patience of 8 epochs. 

All experiments were conducted using a single L40 GPU equipped with 48 GB of GDDR6 memory. Each training routine corresponding to a specific evaluation setup was trained on a dedicated GPU instance. Training EmoForensics took $\approx 3$ hours on a single GPU, whereas Emo-Boost training took $\approx 1$ hour. To ensure computational efficiency and reproducibility, we trained models across different experimental dimensions in parallel, while maintaining consistent resource allocation across runs.

\section{Results}
\label{sec:results}

\begin{table}[h]
\centering
\caption{\textbf{In-Domain Performance Comparison}. Performance comparison between \textbf{Emo-Boost}ed SIMBA, its non–\textbf{Emo-Boost}ed counterpart, and other state-of-the-art multimodal deepfake detectors, evaluated using AUC and AP in the in-domain setting on FakeAVCeleb and DeepSpeak v2. Best results are highlighted in \textbf{bold}, and second-best are \underline{underlined}.}
\label{tab:in_domain_sota}
\resizebox{\columnwidth}{!}{%
\begin{tabular}{cclcl}
\toprule
\multicolumn{1}{l}{} & \multicolumn{2}{c}{FakeAVCeleb} & \multicolumn{2}{c}{DeepSpeak v2} \\
\multicolumn{1}{l}{} & AUC & \multicolumn{1}{c}{AP} & AUC & \multicolumn{1}{c}{AP} \\ \midrule
AVAD~\cite{feng2023self} & 79.16 & 96.09 & 49.88 & 44.65 \\
AVFF\footref{avff}~\cite{oorloff2024avff} & 92.47 & \underline{98.83} & 96.60 & 96.26 \\
SIMBA~\cite{klemt2025deepfakedoctordiagnosingtreating} & \textbf{99.90} & \textbf{99.99} & \textbf{99.79} & \textbf{99.74} \\
\textbf{Emo-Boost}ed SIMBA & \underline{99.89} & \textbf{99.99} & \underline{99.60} & \underline{99.45} \\ \bottomrule
\end{tabular}%
}
\end{table}

\subsection{In-Domain Evaluation}
\label{subsec:in_domain}
We benchmark \textbf{Emo-Boost}ed SIMBA alongside state-of-the-art multimodal deepfake detectors in \textbf{AVAD}~\cite{feng2023self}, \textbf{AVFF}\footnote{We report the results from an \href{https://github.com/JoeLeelyf/OpenAVFF}{unofficial re-implementation} of AVFF.\label{avff}}~\cite{oorloff2024avff}, and the binary variant of \textbf{SIMBA}~\cite{klemt2025deepfakedoctordiagnosingtreating}, on FakeAVCeleb and DeepSpeak v2 under the standard 70\%-30\% split for in-domain evaluation. As shown in~\Cref{tab:in_domain_sota}, across both datasets, Emo-Boosted SIMBA remains competitive with its non-Emo-Boost supported counterpart, which achieves the state-of-the-art performance in this evaluation setup.

\begin{table}[h]
\centering
\caption{\textbf{Summarized Leave-one-out Evaluation.} Comparison of average AUC scores between \textbf{Emo-Boost}ed SIMBA and state-of-the-art multimodal deepfake detectors in the leave-one-out evaluation setup. Emo-Boosted SIMBA achieves the highest average AUC on FakeAVCeleb, improving by 2.1\% over SIMBA, while remaining competitive on DeepSpeak v2. Best values are highlighted in \textbf{bold} and second-best values are \underline{underlined}.}
\label{tab:summary_cross}
\resizebox{\columnwidth}{!}{%
\begin{tabular}{ccc}
\toprule
\multicolumn{1}{l}{} & FakeAVCeleb & DeepSpeak v2 \\ \midrule
AVAD~\cite{feng2023self} & 80.89 & 50.48 \\
AVFF\footref{avff}~\cite{oorloff2024avff} & 86.11 & 93.75 \\
SIMBA~\cite{klemt2025deepfakedoctordiagnosingtreating} & \underline{93.17} & \textbf{95.30} \\
\textbf{Emo-Boost}ed SIMBA & \textbf{95.30} & \underline{95.26} \\ \bottomrule
\end{tabular}%
}
\end{table}

\subsection{Leave-one-out Evaluation}
\label{subsec:cross-man}
\begin{figure*}[t]
    \centering
    \begin{subfigure}[b]{\columnwidth}
        \centering
        \includegraphics[width=\columnwidth]{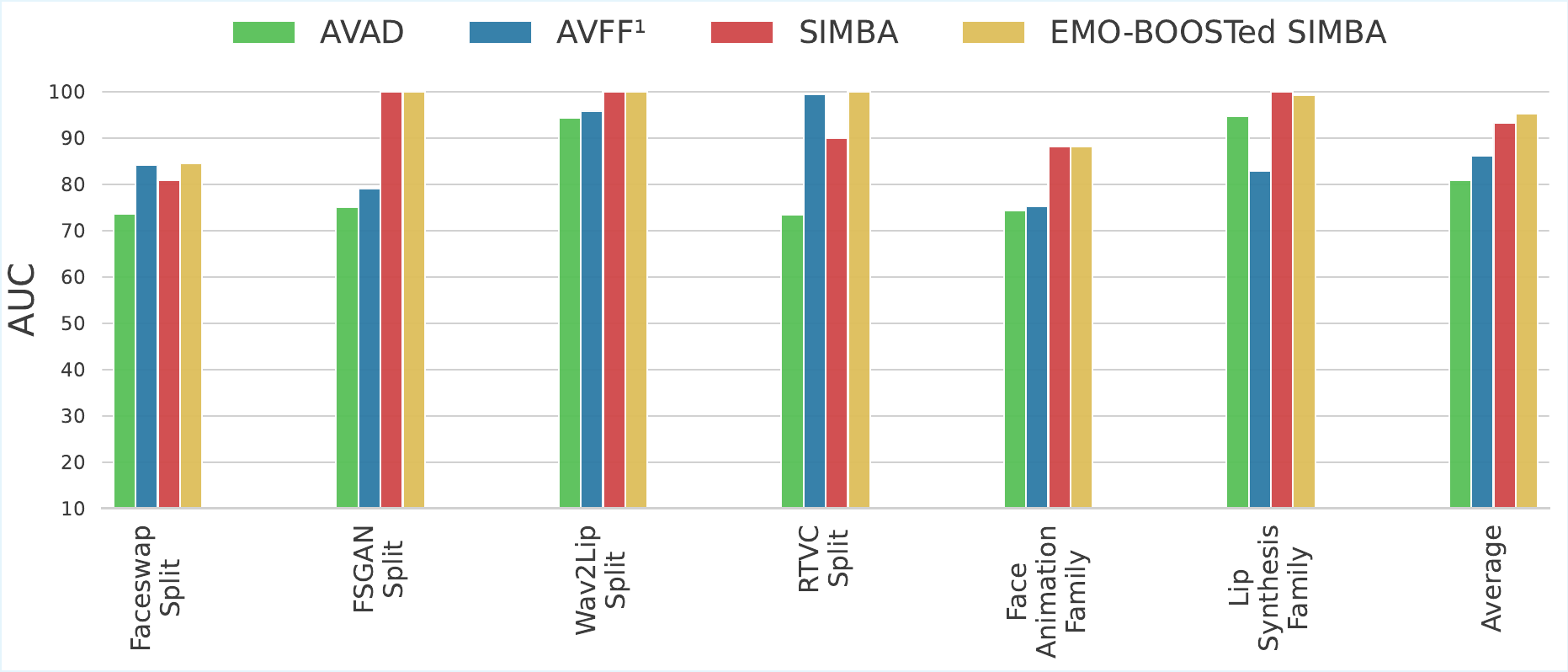}
        \caption{FakeAVCeleb}
        \label{fig:favc_splits}
    \end{subfigure}
    \vspace{0.5em}
    \begin{subfigure}[b]{\columnwidth}
        \centering
        \includegraphics[width=\columnwidth]{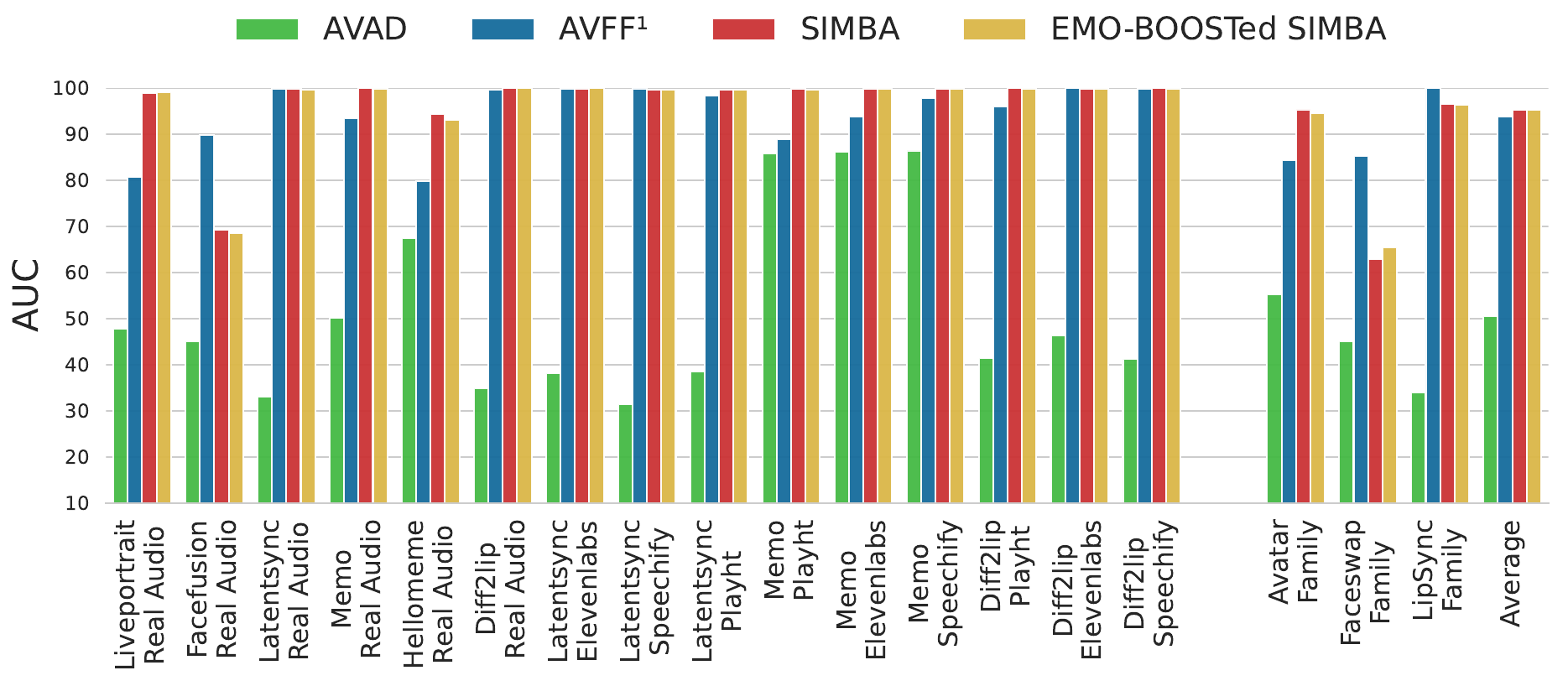}
        \caption{DeepSpeak v2}
        \label{fig:ds2_splits}
    \end{subfigure}
    \caption{\textbf{Leave-one-out Performance Comparison.} Comparison of \textbf{Emo-Boost}ed SIMBA against state-of-the-art multimodal deepfake detectors on FakeAVCeleb and DeepSpeak v2. Emo-Boosted SIMBA achieves a new state-of-the-art average AUC on FakeAVCeleb and improves performance on multiple splits, while remaining competitive across all splits and families on DeepSpeak v2.}
    \label{fig:leave_one_out}
\end{figure*}

We further evaluate \textbf{Emo-Boost}ed SIMBA under the leave-one-out cross-manipulation setting on both FakeAVCeleb and DeepSpeak v2, comparing against the same multimodal baselines as in~\Cref{subsec:in_domain}. As shown in~\Cref{fig:favc_splits}, our method improves over SIMBA on the \textit{Faceswap} split (from 80.87\% to 84.47\% AUC) and the \textit{RealTime Voice Cloning(RTVC)} split (from 89.97\% to 100.0\%) from FakeAVCeleb. The significant improvements reveal that SIMBA and EmoForensics \emph{focus on different signals to detect deepfakes}. Where one model fails to capture the artefact, the other model steps in. In terms of average performance across all splits and families, Emo-Boosted SIMBA achieves the best overall performance on FakeAVCeleb, outperforming the previous best method, SIMBA, by $\approx$2.1\% in AUC. On DeepSpeak v2 (\Cref{fig:ds2_splits}), we do not observe an improvement in the average AUC. However, Emo-Boosted SIMBA remains competitive with SIMBA, disclosing that the simple fusion of both models does not hurt performance, where the low-level focused-SIMBA already achieves high results. We provide a summary of the average leave-one-out performance across datasets in~\Cref{tab:summary_cross}, with the full split-wise results in the Appendix. 

\begin{table}[ht]
\caption{\textbf{EmoForensics Ablation Study.} Comparison between the full EmoForensics pipeline and ablated variants with specific components \textbf{removed} (denoted by \textbf{–}), evaluated on the in-domain AUC on FakeAVCeleb and DeepSpeak v2. Best results are highlighted in \textbf{bold}, and second-best are \underline{underlined}.}
\label{tab:emo-ablation16}
\centering
\resizebox{\columnwidth}{!}{%
\begin{tabular}{lccc}
\toprule
\multicolumn{1}{c}{Methods} & \multicolumn{1}{c}{FakeAVCeleb} & \multicolumn{1}{c}{DeepSpeak v2} & \multicolumn{1}{c}{Modality}\\ \midrule
EmoForensics & \textbf{82.10} & \textbf{65.38} & $\mathcal{AV}$ \\
\hspace{2 mm}– Unimodal Temp Transformers & \underline{80.37} & 63.40 & - \\ 
\hspace{2 mm}– Contrastive Loss & 79.84 & \underline{64.45} & - \\ \midrule
Audio EmoForensics & 73.63 & 53.61 & $\mathcal{A}$ \\
\hspace{2 mm}– Temporal Transformer & 71.48 & 42.42 & - \\ \midrule
Video EmoForensics & 71.15 & 63.01 & $\mathcal{V}$ \\
\hspace{2 mm}– Temporal Transformer & 58.01 & 45.11 & - \\
\bottomrule
\end{tabular}%
}
\end{table}

\begin{figure}[h]
    \centering
    \includegraphics[width=\columnwidth]{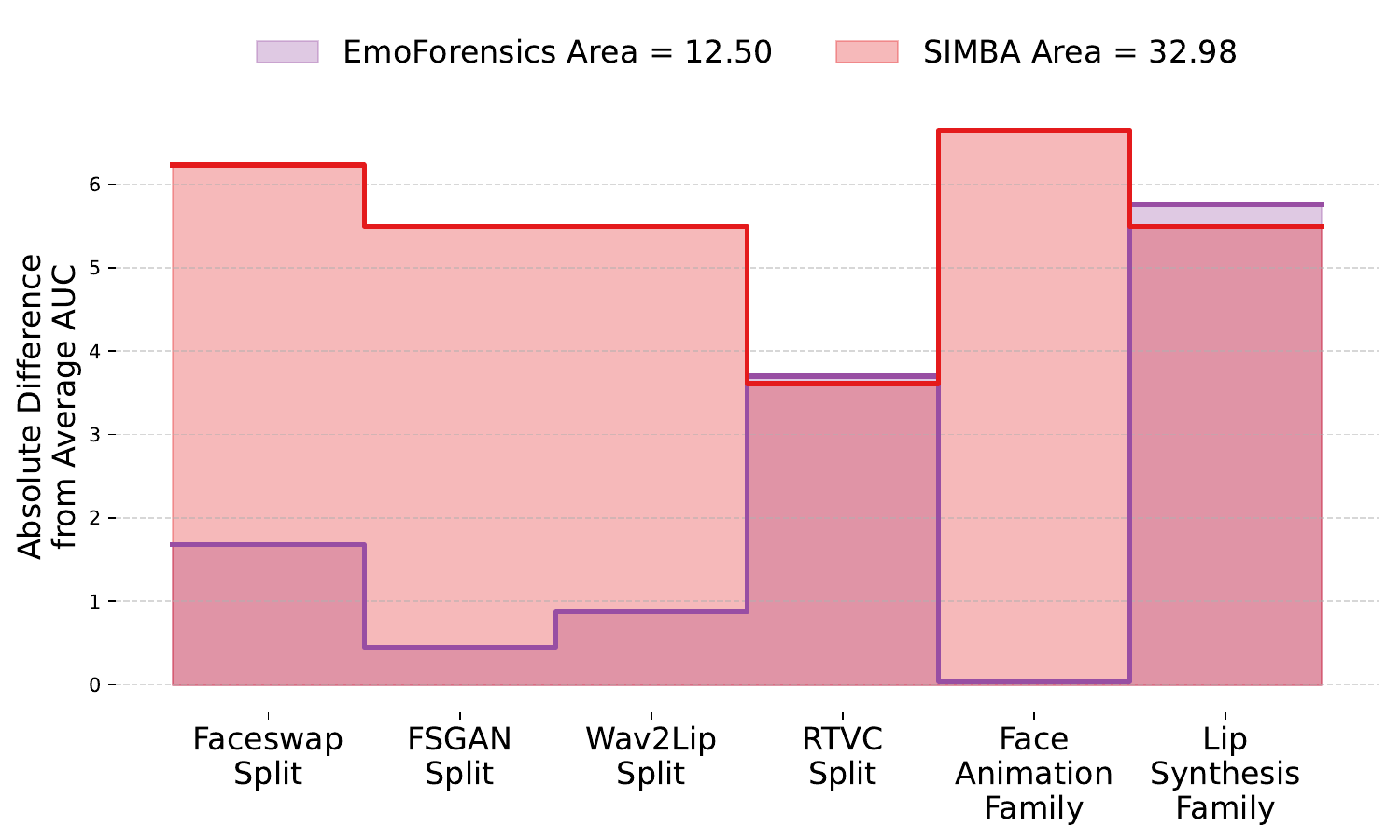}
    \caption{\textbf{Split-wise performance variability of EmoForensics and SIMBA on FakeAVCeleb.} EmoForensics shows smaller fluctuations (Area = 12.50) compared to SIMBA (Area = 32.98), highlighting the \emph{stability} in emotion-based representations from EmoForensics in the cross-manipulation scenario.}
    \label{fig:emotion-stability}
\end{figure}

\subsection{Ablation Study}
\paragraph{EmoForensics} We ablate the EmoForensics pipeline in the in-domain setting by progressively removing key components, including the contrastive loss, modality-specific transformers, and finally both modality transformers, thereby eliminating the model’s ability to capture inter-modal and intra-modal emotion consistency. As shown in~\Cref{tab:emo-ablation16}, removing the unimodal temporal transformers reveals a performance drop of $\approx 2\%$ AUC for both datasets, confirming our hypothesis that modeling temporal consistency unimodally is a helpful signal. To test without the temporal transformers, we average the frame-level visual and audio embeddings $\mathbf{x_v}$ and $\mathbf{x_a}$ across the time dimension, fuse them via element-wise addition, and apply a single-layer trainable classification head. Removing the contrastive loss, on the other hand, results in a performance drop by $2.26\%$ and $0.93\%$ AUC on FakeAVCeleb and DeepSpeak v2, respectively. This drop supports our second hypothesis that learning the emotion consistency across modalities is a useful signal. 

Considering unimodal counterparts, we observe a notable performance drop: on FakeAVCeleb, both settings incur an AUC decrease of $\approx 10\%$, while on DeepSpeak v2 the degradation is asymmetric, with $\approx 2\%$ for Video EmoForensics and $\approx 11\%$ for Audio EmoForensics. Replacing the temporal transformer with a linear layer in the unimodal setting shows that temporal modeling is more critical for video than audio. The Video and Audio EmoForensics pipelines, along with their transformer-free variants, are supervised via modality-specific linear classification heads, which are retained even when temporal transformers are removed. These heads are trained using BCE loss with modality-specific labels. Furthermore, for fairness, unimodal models are evaluated using modality-specific labels as well. In EmoForensics, we use element-wise addition to obtain the joint representation, as it consistently outperforms concatenation and element-wise multiplication across both FakeAVCeleb and DeepSpeak v2 in the in-domain evaluation setup (see Appendix for details).

\paragraph{Why use Emotion} We employ EmoForensics as a complementary pipeline to off-the-shelf RGB- and acoustic-based deepfake detectors due to its observed stability in cross-manipulation evaluation. While SIMBA achieves higher overall AUC scores than EmoForensics, its performance across individual splits shows greater variability relative to its average leave-one-out performance. On some splits, it scores almost to perfection, whereas on others, the generalization performance is significantly lower. To quantify this, we compute the absolute difference between the average and split-wise AUC for each method and visualize the results using a stepwise area plot, shown in~\Cref{fig:emotion-stability}. SIMBA exhibits a larger total area (sum of differences) of 32.98, compared to 12.50 for EmoForensics. We hypothesize that this \textit{emotional stability} contributes to improving the generality of SIMBA under cross-manipulation scenarios. 

\begin{table*}[]
\centering
\caption{\textbf{Emo-Boost Ablation}. Comparison with SIMBA and baseline fusion strategies for integrating EmoForensics with SIMBA, including element-wise feature addition and feature concatenation, evaluated on the curated development set described in~\Cref{sec:results}. Best results are highlighted in \textbf{bold}, and second-best are \underline{underlined}.}
\label{tab:ablation_emoboost}
\resizebox{\textwidth}{!}{%
\begin{tabular}{ccccccccc}
\toprule
\multicolumn{1}{l}{} & \multicolumn{4}{c|}{FakeAVCeleb} & \multicolumn{4}{c}{DeepSpeak v2} \\ \midrule
\multicolumn{1}{l}{} & Overall & \begin{tabular}[c]{@{}c@{}}Faceswap\\ Split\end{tabular} & \begin{tabular}[c]{@{}c@{}}Face Animation\\ Family\end{tabular} & \multicolumn{1}{c|}{Average} & Overall & \begin{tabular}[c]{@{}c@{}}Facefusion\\ Real Split\end{tabular} & \begin{tabular}[c]{@{}c@{}}Lipsync\\ Family\end{tabular} & Average \\ \midrule
SIMBA & 99.74 & 88.28 & 87.86 & 91.96 & \textbf{99.72} & \underline{69.54} & 95.82 & \underline{88.36} \\
SIMBA $\oplus$ EmoForensics & \textbf{99.82} & 89.14 & 88.39 & 92.45 & 99.48 & 67.94 & \underline{95.97} & 87.80 \\
SIMBA $\mid$ EmoForensics & 99.75 & \textbf{90.52} & \underline{88.42} & \underline{92.90} & 99.43 & \textbf{69.59} & 95.92 & 88.31 \\
SIMBA $\odot$ EmoForensics & \underline{99.76} & \underline{89.20} & \textbf{91.01} & \textbf{93.32} & \underline{99.52} & 69.37 & \textbf{96.21} & \textbf{88.37} \\ \bottomrule
\end{tabular}%
}
\end{table*}

\paragraph{Emo-Boost vs Baselines} We compare Emo-Boosted SIMBA against the raw SIMBA and other fusion strategies for combining EmoForensics and SIMBA features in~\Cref{tab:ablation_emoboost} in the cross-manipulation setting. As possible fusion strategies, we consider element-wise addition ($\oplus$) and feature concatenation ($\mid$), while Emo-Boosted SIMBA is denoted as element-wise product ($\odot$). Experiments are conducted on a curated development set. The overall performance follows the conventional in-domain setup, but the cross-manipulation split results are on the \textit{val-test} split introduced in~\Cref{sec:exp_setup}. Although several fusion strategies improve performance on certain splits, the element-wise product consistently achieves the highest average AUC, improving by $\approx$1.3\% on FakeAVCeleb over SIMBA and showing a marginal improvement on DeepSpeak v2. 
\section{Conclusion, Limitations and Future Work}
\label{sec:concs}

In this paper, we propose \textbf{Emo-Boost}, an emotion-augmented multimodal deepfake detection framework that improves cross-manipulation generalization performance. The framework augments signal extracted from a frozen off-the-shelf RGB and acoustic-based deepfake detector, which captures low-level information, with our emotion-based multimodal deepfake detector \textbf{EmoForensics}, that focuses on high-level semantic cues. EmoForensics models the emotion signal from a frozen video and audio emotion encoder along two dimensions. First, intra-modal temporal consistency in emotions is considered via two temporal transformers. Second, emotion consistency across the visual and audio modality is modeled via contrastive loss. We observe that the low-level deepfake detector and EmoForensics capture different signals in deepfakes. Thus, fusing them in Emo-Boost via simple element-wise multiplication improves cross-manipulation performance on FakeAVCeleb by 2.1\% in AUC, while remaining competitive with state-of-the-art multimodal deepfake detectors in in-domain evaluation and in cross-manipulation performance on DeepSpeak v2.

\paragraph{Limitations} The primary limitation of our work is the weaker performance of EmoForensics in a standalone setting. Additionally, we observe little to no improvement in performance on DeepSpeak v2, which we attribute to the differences in dataset construction. While FakeAVCeleb consists of videos collected in the wild, DeepSpeak v2 follows a scripted recording protocol, thus restricting the natural expressivity of emotions available for the model to capture. This is further supported by our ablation study, where transformer-free unimodal variants achieve stronger performance on FakeAVCeleb than on DeepSpeak v2, indicating that emotional cues are less pronounced in DeepSpeak v2 and thus provide less signal. We also acknowledge that our framework may inherit potential biases present in the datasets used for training.

\paragraph{Future Work} From our perspective, promising future avenues are: (i) exploring other high-level semantic cues and their combination for deepfake detection. As we observed that our models capture different signals, other cues like mimic or movement could be combined with detectors that focus on pixel-level artefacts. (ii) the creation of in-the-wild deepfake datasets with recent manipulation techniques that represent how humans interact in the real world. This would allow the high-level focused approaches to capture signals present in natural human behavior, but not in a staged recording environment.

\paragraph{Acknowledgements} We gratefully acknowledge support from the Konrad Zuse School of Excellence in Learning and Intelligent Systems (ELIZA) during the Master’s studies of one of the authors, as part of the ``Konrad Zuse Schools of Excellence in Artificial Intelligence" programme financed by the German Academic Exchange Service~(DAAD) and the German Federal Ministry of Education and Research~(BMBF).\\
We gratefully acknowledge support from the hessian.AI Service Center (funded by the Federal Ministry of Research, Technology and Space, BMFTR, grant no. 16IS22091) and the hessian.AI Innovation Lab (funded by the Hessian Ministry for Digital Strategy and Innovation, grant no. S-DIW04/0013/003).

{
    \small
    \bibliographystyle{ieeenat_fullname}
    \bibliography{main}
}

\clearpage
\setcounter{page}{1}
\maketitlesupplementary

We structure the supplementary materials as follows: first,~\Cref{sec:dataset_supp} provides details on our employed datasets, FakeAVCeleb and DeepSpeak v2, used for training, validation, and testing our methods across different evaluation scenarios and for benchmarking against state-of-the-art methods. Further, in~\Cref{sec:more_results}, we provide the detailed tabular representations of the comparison of our framework in the leave-one-out evaluation scenario on both datasets, along with the tabular comparison of different fusion methods implemented in EmoForensics in the in-domain evaluation on FakeAVCeleb and DeepSpeak v2, and stability in performance between EmoForensics and SIMBA on the leave-one-out evaluation setup on FakeAVCeleb. 

\section{Details on Dataset}
\label{sec:dataset_supp}

In this section, we provide some details on how we split and utilise the FakeAVCeleb~\cite{khalid2021fakeavceleb} and DeepSpeak v2~\cite{barrington2024deepspeak} datasets for benchmarking our proposed framework Emo-Boost. We create a validation split for both datasets, which is used for learning rate scheduling and early stopping. Furthermore, we also validate our design choices for our methods based on the performance on this split. Method and Family splits are constructed following the setup provided by~\cite{klemt2025deepfakedoctordiagnosingtreating}. As described in~\Cref{sec:exp_setup}, we introduce a new \textit{val-test} split for the purpose of validating our design choices in the cross-manipulation evaluation scenario. For every Method and Family split, the val-test split is constructed by uniformly sampling 20\% of both real and fake videos from the corresponding test set. We use this split only for the purpose of model selection and hyperparameter tuning. The model does not observe this split during training, and only observes the in-domain train and validation split, the latter being used for learning rate scheduling and early stopping. For reporting our results, we use the entire test split as provided, ie, by adding the newly created val-test split back to the test split.

\subsection{FakeAVCeleb}
We construct the train, validation, and test splits in FakeAVCeleb following the same mechanism as in~\cite{klemt2025deepfakedoctordiagnosingtreating} in the ratio of 60\%, 10\%, and 30\%, respectively, on the basis of provided identity annotations. Similarly, following~\cite{klemt2025deepfakedoctordiagnosingtreating}, we use four Method splits (Faceswap, FSGAN, Wav2Lip, and RealTime Voice Cloning) and two Family splits (Face Animation and Lip Synthesis) for our cross-manipulation evaluation setup. In these Method and Family splits, we introduce the val-test split as described before. For example, for the Faceswap method split, while the train and val splits do not contain any fake video manipulated using the Faceswap method, the test set contains 929 such videos, along with 150 real videos. To construct our val-test split, we sample 20\% of each class. Therefore, the final val-test split for leave-one-out evaluation of Faceswap contains 30 real videos and 186 fake videos. These videos are then added back to the test set when obtaining the final results. The train, val, and test splits for the standard in-domain evaluation setup contain \textbf{12,935}, \textbf{2,176}, and \textbf{6,455} samples, respectively.

\subsection{DeepSpeak v2}
For the construction of the validation split in DeepSpeak v2, we use 20\% of the training data while the test set remains untouched. The Method and Family splits, including the internal val-test split, are constructed in the same manner as in FakeAVCeleb, following~\cite{klemt2025deepfakedoctordiagnosingtreating}. In DeepSpeak v2, we use 15 Method Splits and 3 Family Splits for the leave-one-out evaluation setup. The train, validation, and test splits of the standard in-domain evaluation setup for DeepSpeak v2 comprise \textbf{10,645}, \textbf{2,661}, and \textbf{3,279} samples, respectively.

\section{Further Results}
\label{sec:more_results}

\subsection{Leave-one-out Evaluation}
As described in~\Cref{subsec:cross-man}, we observe motivating results from our proposed framework in the leave-one-out evaluation setup on FakeAVCeleb. We present the detailed performance and its comparison with other state-of-the-art multimodal deepfake detectors in~\Cref{tab:cross_manipulation_favc}. \textbf{Emo-Boost}ed SIMBA achieves the highest AUC across all splits except for the Face Animation Family, where it is the second best, and also attains the best average performance.

Similarly, we provide the detailed comparison of Emo-Boosted SIMBA against the other deepfake detectors in the same evaluation setup on DeepSpeak v2 in~\Cref{tab:cross_manipulation_ds2}. While \textbf{Emo-Boost}ed SIMBA shows minor improvements on splits such as \textit{LivePortrait Real}, \textit{LatentSync ElevenLabs}, and \textit{LatentSync Speechify}, its average performance remains slightly below that of SIMBA.

\subsection{Further Ablations}

\paragraph{EmoForensics} We provide the detailed comparison of different fusion strategies used in the EmoForensics framework to obtain the joint representation on the in-domain evaluation setup on both FakeAVCeleb and DeepSpeak v2 in~\Cref{emoforensicsFusion}. 

\begin{table}[!ht]
    \centering
    \caption{\textbf{Fusion Strategy Ablation.} Comparison of different fusion strategies within the EmoForensics framework on the in-domain evaluation setting for FakeAVCeleb and DeepSpeak v2, reported in terms of AUC score. Element-wise addition yields the best performance across both datasets. Best results are highlighted in \textbf{bold}, and second-best are \underline{underlined}.}
    \resizebox{\linewidth}{!}{%
    \begin{tabular}{lcc}
    \toprule
        EmoForensics & FakeAVCeleb & DeepSpeak v2 \\ \midrule
         w/ Element-wise Addition & \textbf{82.10} & \textbf{65.38} \\ 
         w/ Concatenation & 80.85 & \underline{64.52} \\ 
         w/ Element-wise Product & \underline{81.80} & 58.46 \\ \bottomrule
    \end{tabular}%
    }
    \label{emoforensicsFusion}
\end{table}

\paragraph{Why use Emotion} We provide the detailed comparison of EmoForensics and SIMBA on the leave-one-out evaluation setup on FakeAVCeleb in~\Cref{tab:emotion_stability}. We highlight fluctuations in performance across individual Methods and Family splits relative to each model's average AUC in this evaluation setup. 

\begin{table*}[t]
\centering
\caption{\textbf{Cross-manipulation performance comparison on FakeAVCeleb.} Performances are given as AUC. Best results are highlighted in \textbf{bold}, while the second-best are \underline{underlined}.}
\label{tab:cross_manipulation_favc}
\begin{tabular}{cccccccc}
\toprule
\multicolumn{1}{l}{} & \begin{tabular}[c]{@{}c@{}}Faceswap\\ Split\end{tabular} & \begin{tabular}[c]{@{}c@{}}FSGAN\\ Split\end{tabular} & \begin{tabular}[c]{@{}c@{}}Wav2Lip\\ Split\end{tabular} & \begin{tabular}[c]{@{}c@{}}RTVC\\ Split\end{tabular} & \begin{tabular}[c]{@{}c@{}}Face\\ Animation\\ Family\end{tabular} & \begin{tabular}[c]{@{}c@{}}Lip\\ Synthesis\\ Family\end{tabular} & Average \\ \midrule
AVAD~\cite{feng2023self} & 73.63 & 75.01 & 94.27 & 73.34 & 74.32 & 94.75 & 80.89 \\
AVFF\footref{avff}~\cite{oorloff2024avff} & \underline{84.07} & \underline{79.14} & \underline{95.86} & \underline{99.38} & 75.30 & 82.91 & 86.11 \\
SIMBA~\cite{klemt2025deepfakedoctordiagnosingtreating} & 80.87  & \textbf{100.00} & \textbf{99.98} & 89.97 & \textbf{88.21} & \textbf{100.00} & \underline{93.17} \\
EMO-BOOSTed SIMBA & \textbf{84.47} & \textbf{100.00} & \textbf{99.98} & \textbf{100.00} & \underline{88.11} & \underline{99.23} & \textbf{95.30} \\ \bottomrule
\end{tabular}%
\end{table*}

\begin{table*}[t]
\centering
\caption{\textbf{Cross-manipulation method split performance on DeepSpeak v2.} Performances are given as AUC. Both SIMBA and Emo-Boosted SIMBA outperform the other deepfake detectors in terms of average AUC. Best results are highlighted in \textbf{bold}, and second-best are \underline{underlined}.}
\label{tab:cross_manipulation_ds2}
\resizebox{\textwidth}{!}{%
\begin{tabular}{cccccccccccccccc}
\hline
\multicolumn{1}{l}{}                                   & \begin{tabular}[c]{@{}c@{}}Liveportrait\\ Real\\ Audio\end{tabular} & \begin{tabular}[c]{@{}c@{}}Facefusion\\ Real\\ Audio\end{tabular} & \begin{tabular}[c]{@{}c@{}}Latentsync\\ Real\\ Audio\end{tabular} & \begin{tabular}[c]{@{}c@{}}Memo\\ Real\\ Audio\end{tabular} & \begin{tabular}[c]{@{}c@{}}Hellomeme \\ Real\\ Audio\end{tabular} & \begin{tabular}[c]{@{}c@{}}Diff2lip\\ Real\\ Audio\end{tabular} & \begin{tabular}[c]{@{}c@{}}Latentsync\\ Elevenlabs\end{tabular} & \begin{tabular}[c]{@{}c@{}}Latentsync\\ Speechify\end{tabular} & \begin{tabular}[c]{@{}c@{}}Latentsync\\ Playht\end{tabular} & \begin{tabular}[c]{@{}c@{}}Memo\\ Playht\end{tabular} & \begin{tabular}[c]{@{}c@{}}Memo\\ Elevenlabs\end{tabular} & \begin{tabular}[c]{@{}c@{}}Memo\\ Speechify\end{tabular} & \begin{tabular}[c]{@{}c@{}}Diff2lip\\ Playht\end{tabular} & \begin{tabular}[c]{@{}c@{}}Diff2lip\\ Elevenlabs\end{tabular} & \begin{tabular}[c]{@{}c@{}}Diff2lip\\ Speechify\end{tabular} \\ \hline
AVAD~\cite{feng2023self}                               & 47.81                                                               & 45.10                                                             & 33.05                                                             & 50.25                                                       & 67.52                                                             & 34.95                                                           & 38.27                                                           & 31.54                                                          & 38.49                                                       & 85.76                                                 & 86.20                                                     & 86.46                                                    & 41.42                                                     & 46.35                                                         & 41.19                                                        \\
AVFF\footref{avff}~\cite{oorloff2024avff}              & 80.80                                                               & \textbf{89.76}                                                    & \underline{99.81}                                                   & 93.38                                                       & 79.88                                                             & 99.71                                                           & \underline{99.85}                                                 & \textbf{99.91}                                                 & 98.33                                                       & 88.94                                                 & 93.82                                                     & 97.76                                                    & 96.07                                                     & \textbf{99.99}                                                & \underline{99.89}                                              \\
SIMBA~\cite{klemt2025deepfakedoctordiagnosingtreating} & \underline{98.94}                                                     & \underline{69.19}                                                   & \textbf{99.85}                                                    & \textbf{99.95}                                              & \textbf{94.36}                                                    & \textbf{100.00}                                                 & \underline{99.85}                                                 & 99.66                                                          & \textbf{99.61}                                              & \textbf{99.77}                                        & \textbf{99.87}                                            & \textbf{99.78}                                           & \textbf{99.93}                                            & 99.74                                                         & \textbf{100.00}                                              \\
EMO-BOOSTed SIMBA                                      & \textbf{99.03}                                                      & 68.58                                                             & 99.60                                                             & \underline{99.90}                                             & \underline{93.17}                                                   & \underline{99.96}                                                 & \textbf{99.92}                                                  & \underline{99.71}                                                & \underline{99.57}                                             & \underline{99.64}                                       & \underline{99.74}                                           & \underline{99.75}                                          & \underline{99.87}                                           & \underline{99.90}                                               & 99.79                                                        \\ \hline
\end{tabular}%
}
\end{table*}

\begin{table*}[t]
\centering
\caption{\textbf{Cross-manipulation family split performance on DeepSpeak v2.} Family split performances are shown next to the average performance that is calculated across the method(\Cref{tab:cross_manipulation_ds2}) and family splits. Performances are given as AUC. Best results are highlighted in \textbf{bold}, and second-best are \underline{underlined}.}
\label{tab:cross_manipulation_ds2_fam}
\begin{tabular}{ccccc}
\hline
\multicolumn{1}{l}{}                                   & \begin{tabular}[c]{@{}c@{}}Avatar\\ Family\end{tabular} & \begin{tabular}[c]{@{}c@{}}Faceswap\\ Family\end{tabular} & \begin{tabular}[c]{@{}c@{}}LipSync\\ Family\end{tabular} & Average         \\ \hline
AVAD~\cite{feng2023self}                               & 55.19                                                   & 45.10                                                     & 34.00                                                    & 50.48           \\
AVFF\footref{avff}~\cite{oorloff2024avff}              & 84.36                                                   & \textbf{85.28}                                            & \textbf{99.97}                                           & 93.75           \\
SIMBA~\cite{klemt2025deepfakedoctordiagnosingtreating} & \textbf{95.37}                                          & 62.87                                                     & \underline{96.59}                                          & \textbf{95.30}  \\
EMO-BOOSTed SIMBA                                      & \underline{94.58}                                         & \underline{65.48}                                           & 96.45                                                    & \underline{95.26} \\ \hline
\end{tabular}%
\end{table*}

\begin{table*}[t]
\centering
\caption{\textbf{Stability Ablation}. Comparison of performance variation across manipulation splits in the cross-manipulation evaluation setup for EmoForensics and SIMBA. Values in brackets indicate the difference from the average AUC across all manipulation and family splits for a given method. Red values denote a drop in performance relative to the average AUC, while green values indicate an increase.}
\label{tab:emotion_stability}
\resizebox{\textwidth}{!}{%
\begin{tabular}{cccccccc}
\toprule
\multicolumn{1}{l}{} & \begin{tabular}[c]{@{}c@{}}Faceswap\\ Split\end{tabular} & \begin{tabular}[c]{@{}c@{}}FSGAN\\ Split\end{tabular} & \begin{tabular}[c]{@{}c@{}}Wav2Lip\\ Split\end{tabular} & \begin{tabular}[c]{@{}c@{}}RTVC\\ Split\end{tabular} & \begin{tabular}[c]{@{}c@{}}Face Animation\\ Family\end{tabular} & \begin{tabular}[c]{@{}c@{}}Lip Synthesis\\ Family\end{tabular} & Average \\ \midrule
EmoForensics &  70.98(\textcolor{ForestGreen}{+1.68}) & 68.85(\textcolor{red}{-0.45}) & 70.17(\textcolor{ForestGreen}{+0.87}) & 73.00(\textcolor{ForestGreen}{+3.70}) & 69.26(\textcolor{red}{-0.04}) & 63.54(\textcolor{red}{-5.76}) & 69.30 \\
SIMBA & 88.28(\textcolor{red}{-6.23}) & 100.00(\textcolor{ForestGreen}{+5.50}) & 100.00(\textcolor{ForestGreen}{+5.50}) & 90.89(\textcolor{red}{-3.61}) & 87.86(\textcolor{red}{-6.65}) & 100.00(\textcolor{ForestGreen}{+5.50}) & 94.50 \\ \bottomrule
\end{tabular}%
}
\end{table*}

\end{document}